\documentclass[twoside,11pt]{article}
\usepackage{paralist,amsmath, amssymb}
\usepackage{algorithm,algorithmic}
 \usepackage{graphicx}
 \usepackage{epstopdf}

\def \E {\mathrm{E}}

\def \D {\mathcal{D}}

\def \w {\mathbf{w}}
\def \R {\mathbb{R}}

\def \p {\mathbf{p}}

\def \m {\mathbf{m}}

\newtheorem{ass}{Assumption}
\newtheorem{defn}{Defination}

\DeclareMathOperator*{\argmin}{argmin}
\DeclareMathOperator*{\argmax}{argmax}
\usepackage{microtype}
\usepackage{graphicx}
\usepackage{subfigure}
\usepackage{booktabs} 

\usepackage{hyperref}

\usepackage{jmlr2e}

\begin{document}

\title{Online Convex Optimization with  Continuous \\ Switching Constraint}

\author{\name Guanghui Wang \email wanggh@lamda.nju.edu.cn\\
\name Yuanyu Wan \email wanyy@lamda.nju.edu.cn\\
       \addr National Key Laboratory for Novel Software Technology  Nanjing University, Nanjing 210023, China\\
       \name Tianbao Yang\email tianbao-yang@uiowa.edu\\
       \addr Department of Computer Science, the University of Iowa, Iowa City, IA 52242, USA\\
        \name Lijun Zhang \email zhanglj@lamda.nju.edu.cn\\
       \addr National Key Laboratory for Novel Software Technology   Nanjing University, Nanjing 210023, China}

\maketitle

\begin{abstract}
 
\noindent In many sequential decision making applications, the change of decision would bring an additional cost, such as the wear-and-tear cost associated with changing server status. To control the switching cost, we introduce the problem of online convex optimization with continuous switching constraint, where the goal is to achieve a small regret given a budget on the \emph{overall} switching cost. We first investigate the hardness of the problem, and provide a lower bound of order $\Omega(\sqrt{T})$ when the switching cost budget $S=\Omega(\sqrt{T})$, and $\Omega(\min\{\frac{T}{S},T\})$ when $S=O(\sqrt{T})$, where $T$ is the time horizon. The essential idea is to carefully design an adaptive adversary, who can adjust the loss function according to the cumulative switching cost of the player incurred so far based on the orthogonal technique. We then develop a simple gradient-based algorithm which enjoys the minimax optimal regret bound. Finally, we show that, for strongly convex functions, the regret bound can be improved to $O(\log T)$ for $S=\Omega(\log T)$, and $O(\min\{T/\exp(S)+S,T\})$ for $S=O(\log T)$.





\end{abstract}

\section{Introduction}
Online convex optimization (OCO) is a fundamental framework for studying sequential decision making problems \citep{Online:suvery}. Its protocol can be seen as a game between a player and an adversary: In each round $t$, firstly, the player selects an action $\w_t$ from a convex set $\D\subseteq\R^d$. After submitting the answer, a loss function $f_t:\mathcal{D} \mapsto \R$ is revealed, and the player suffers a loss $f_t(\w_t)$. The goal is to minimize the regret:
\begin{equation}
\label{defn:dynamic-regret}
R=\sum_{t=1}^Tf_t(\w_t)-\min\limits_{\w\in\D}\sum_{t=1}^Tf_t(\w),
\end{equation}
which is the difference between the cumulative loss of the player and that of the best action in hindsight. 

Over the past decades, the problem of OCO has been extensively studied, yielding various algorithms and theoretical guarantees \citep{Intro:Online:Convex,orabona2019modern}. However, most of the existing approaches allow the player to switch her action~\emph{freely}~during the learning process. As a result, these methods become unsuitable for many real-life scenarios, such as the online shortest paths problem \citep{koolen2010hedging}, and portfolio management \citep{dekel2014bandits}, where the switching of actions brings extra cost, and the budget for the overall switching cost is strictly constrained. To address this problem, recent advances in OCO introduced the switching-constrained problem \citep{altschuler2018online,chen2019minimax}, where a hard constraint is imposed to the \emph{number} of the player's action shifts, i.e.,
\begin{equation}
\label{intro2}
\sum_{t=2}^T\{\w_{t}\not=\w_{t-1}\}\leq K,
\end{equation}
and the goal is to minimize regret under a fixed budget $K$. For this problem, \citet{chen2019minimax} have shown that, given any $K$, we could precisely control the the overall switching cost in \eqref{intro2}, while achieving a minimax regret bound of order $\Theta({T}/{\sqrt{K}})$.

One limitation of \eqref{intro2} is that it treats different amounts of changes between $\w_{t-1}$ and $\w_{t}$ equally, since the \emph{binary} function is used as the penalty for action shifts. However, as observed by many practical applications, e.g.,  thermal management \citep{zanini2010online},
video streaming \citep{joseph2012jointly} and multi-timescale control \citep{goel2017thinking}, the price  paid for large and small action changes are not the same. Specifically, for these scenarios, the switching cost between two consecutive rounds is typically characterized by a $\ell_2$-norm function, i.e., $\|\w_t-\w_{t-1}\|$. Motivated by this observation, in this paper, we introduce a novel OCO setting, named OCO with continuous switching constraint (OCO-CSC), where the player needs to choose actions under a hard constraint on the \emph{overall $\ell_2$-norm switching cost}, i.e.,
\begin{equation}
\label{eqn:intro:ours}
\sum_{t=2}^T\|\w_{t}-\w_{t-1}\|\leq S,
\end{equation}
where $S$ is a budget given by the environment. The main advantage of OCO-CSC is that, equipped with \eqref{eqn:intro:ours}, we could have a more delicate control on the overall switching cost  compared to the binary constraint in \eqref{intro2}. 

For the proposed problem, we firstly observe that, an $O(T/\sqrt{S})$ regret bound can be achieved by using the method proposed for the switching-constrained OCO \citep{chen2019minimax} under a proper configuration of $K$. However, this bound is not tight, since there is a large gap from the lower bound established in this paper. Specifically, we provide a lower bound of order $\Omega(\sqrt{T})$ when $S=\Omega(\sqrt{T})$, and $\Omega(\min\{\frac{T}{S},T\})$ when $S=O(\sqrt{T})$. The basic framework for constructing the lower bound follows the classical linear game \citep{Minimax:Online}, while we adopt a novel mini-batch policy for the adversary, which allows it to adaptively change the loss function according to the player's cumulative switching costs. Furthermore, we prove that the classical online gradient descent (OGD) with an appropriately chosen step size is able to obtain the matching upper bound. These results demonstrate that there is a \emph{phase transition} phenomenon between large and small switching budget regimes, which is in sharp contrast to the switching-constrained setting, where the minimax bound always decreases with $\Theta(1/\sqrt{K})$. Finally, we propose a variant of OGD for $\lambda$-strongly convex functions, which can achieve an $O(\log T)$ regret bound when $S=\Omega(\log T)$, and an $O(T/\exp(S)+S)$ regret bound when $S=O(\log T)$.    

\section{Related Work}
In this section, we briefly review related work on online convex optimization.
\subsection{Classical OCO}
The framework of OCO is established by the seminal work of \citet{zinkevich-2003-online}. For general convex functions, \citet{zinkevich-2003-online} shows that online gradient descent (OGD) with step size on the order of $O(1/\sqrt{t})$ enjoys an $O(\sqrt{T})$ regret bound. For
$\lambda$-strongly convex functions, \citet{Hazan:2007:log} prove that OGD with step size of order $O(1/[\lambda t])$ achieves an $O(\log T)$ regret bound. Both bounds have been proved to be minimax optimal \citep{Minimax:Online}. For exponentially concave functions, the state-of-the-art algorithm is online Newton step (ONS), which enjoys an $O(d\log T)$ regret bound, where $d$ is the dimensionality.

\subsection{Switching-constrained OCO}
\label{2.3}
One relted line of research is the switching-constrained setting, where the player is only allowed to change her action no more than $K$ times. This setting has been studied in various online learning scenarios, such as prediction with expert advice \citep{altschuler2018online} and  bandits problems \citep{switchbandits2019,dong2020multinomial,ruan2020linear}. In this paper, we focus on online convex optimization. \citet{jaghargh2019consistent} firstly consider this problem, and develop a novel online algorithm based on the Poison Process, which can achieve an expected regret of order $O({T^{3/2}}/{\E[K]})$ for any given expected switching budget $\E[K]$. Therefore, the regret will become sublinear for $\E{[K]}=o(\sqrt{T})$. Later, \citet{chen2019minimax} propose a variant of the classical OGD based on the mini-batch approach, which enjoys an $O(T/\sqrt{K})$ regret bound for any given budget $K$.  They also prove that this result is minimax optimal by  establishing a matching $\Omega(T/\sqrt{K})$ lower bound. We note that, when the action set is bounded (i.e., $\max_{\w_1,\w_2\in \D}\|\w_1-\w_2\|\leq D$), since
$$\sum_{t=2}^T\|\w_{t}-\w_{t-1}\|\leq DK,$$
we could set $K=\lfloor {S}/{D} \rfloor$ to satisfy \eqref{eqn:intro:ours} and immediately obtain an $O(T/\sqrt{S})$ regret for OCO-CSC, but there is still a large gap from the lower bound we provide in this paper. 

\subsection{OCO with Ramp Constraints}
Another related setting is OCO with ramp constraints, which is studied by \citet{badiei2015online}. In this setting, at each round, the player must choose an action satisfying the following inequality: 
\begin{equation}
\label{eqn:ramp}
|w_{t,i}-w_{t-1,i}|\leq X_i,
\end{equation}
where $w_{t,i}$ denotes the $i$-th dimension of $\w_t$, and $X_i$ is a constant factor. The constraint in \eqref{eqn:ramp} limits the player's action switching in a \emph{per-round} and \emph{per-dimension} level. This is very different from the constraint we proposed in \eqref{eqn:intro:ours}, which mainly focus on the \emph{long-term} and \emph{overall} switching cost. Moreover, we note that, \citet{badiei2015online} assume the player could get access to a sequence of \emph{future} loss functions before choosing $\w_t$, while in this paper we follow the classical OCO framework in which the player can only make use of the historical data.
\subsection{OCO with Long-term Constraints}
Our proposed problem is also related to OCO with long-term constraints \citep{mahdavi2012trading,jenatton2016adaptive,yu2017online}, where the action set is written as $m$ convex constraints, i.e., 
\begin{equation}
\label{eqn:related work}
	\D=\{\w\in\R^d: g_i(\w)\leq 0, i\in [m]\},
\end{equation}
 and we only require these constraints to be satisfied in the long term, i.e., $\sum_{t=1}^Tg_i(\w_t)\leq 0, i\in[m]$. The goal is to minimize regret while keeping $\sum_{t=1}^Tg_i(\w_t)$ small. We note that, in this setting, the action set is \emph{expressed} by the constraint, which is in contrast to OCO-CSC, where the constraint and the decision set $\D$ are independent. Moreover, the constraint in OCO-CSC is time-variant and decided by the historical decisions, while 
 the constraint in \eqref{eqn:related work} is static (or stochastic, considered by \citeauthor{yu2017online}, \citeyear{yu2017online}). 
  Recently, several work start to investigate OCO with long-term and time-variant constraints, but this task  is proved to be impossible in general \citep{mannor2009online}. Therefore, existing studies have to consider more restricted settings, such as weaker definitions of regret \citep{neely2017online,liakopoulos2019cautious,yi2020distributed,valls2020online}. 

\subsection{Smoothed OCO}
The problem of smoothed OCO is originally proposed in the  
dynamic right-sizing for power-proportional data centers \citep{lin2012dynamic}, and has received great research interests during the past decade \citep{lin2012online,bansal20152,antoniadis2017tight,chen2018smoothed,goel2019beyond}. In smoothed OCO, at each round, the learner will incur a \emph{hitting} cost $f_t(\cdot)$ as well as a \emph{switching} cost $\|\w_t-\w_{t-1}\|$, and the goal is to minimize dynamic regret \citep{zinkevich-2003-online} or competitive ratio \citep{borodin2005online} with respect to $f_t(\w_t)+\|\w_t-\w_{t-1}\|$. 
This setting is significantly different from OCO-CSC, where the goal is to minimize regret with respect to $f_t(\cdot)$, and the overall switching cost is limited by a given budget. Additionally, we note that, similar to \citet{badiei2015online}, studies for the smoothed OCO  typically assume the player could see $f_t(\cdot)$ or sometimes a window of future loss functions \citep{chen2015online,chen2016using,li2018using} before choosing $\w_t$. By contrast, in OCO-CSC the player can not obtain these additional information. 

\section{Main Results}
In this section, we present the algorithms and theoretical guarantees for OCO-CSC.
Before proceeding to the details, following previous work, we introduce some standard definitions \citep{Convex-Optimization} and assumptions \citep{Minimax:Online}.
\begin{defn}
A function $f:\mathcal{D}\mapsto \mathbb{R}$ is convex if  $\forall \w_1, \w_2\in\D,$ \emph{
\begin{equation}
\begin{split}
\label{defn:convex}
f(\w_1)\geq f(\w_2)+\nabla f(\w_2)^{\top}(\w_1-\w_2).
\end{split}
\end{equation}}
\end{defn}
\begin{defn}
\label{defn:stconvex}
A function $f:\mathcal{D}\mapsto \mathbb{R}$ is $\lambda$-strongly convex if  \emph{$\forall \w_1, \w_2\in\mathcal{D}$,
\begin{equation}
\begin{split}
\label{eq:strongly-convex}
f(\w_1)\geq & f(\w_2)+ \nabla f(\w_2)^{\top}(\w_1-\w_2) +\frac{\lambda}{2}\|\w_1-\w_2\|^2.
\end{split}
\end{equation}}
\end{defn}
\begin{ass}\label{ass:1} The action set $\D$ is a $d$-dimensional ball of radius $\frac{D}{2}$, i.e., $$\D=\left\{\w\bigg|\w\in\R^d, \|\w\|\leq \frac{D}{2}\right\}.$$
\end{ass}
\begin{ass}\label{ass:2} The gradients of all the online functions are bounded by $G$, i.e.,
\begin{equation}\label{eqn:gradient}
\max_{\w \in \D}\|\nabla f_t(\w)\| \leq G, \ \forall t \in[T].
\end{equation}
\end{ass}
\subsection{Lower Bound for Convex Functions}
\begin{algorithm}
   \caption{Adversary's Policy}
   \label{alg:AP}
\begin{algorithmic}[1]
\STATE $\tau=1$, $l_{\tau}=1$
\STATE Observe the player's action $\w_1$
\STATE Choose $\m_1$ such that $\m_1^{\top}\w_1=0$
   \FOR{$t=2$ {\bfseries to} $T$}
   \STATE Observe the player's action $\w_t$
   \IF{$\sum^t_{j=l_{\tau}+1}\|\w_j-\w_{j-1} \|\leq \frac{c}{S}$}
   \STATE Choose $\m_t=\m_{t-1}$
   \ELSE
   \STATE Choose $\m_t$ such that $\m_t^{\top}\w_t\geq0$ and $\m_t^{\top}\left(\sum_{j=1}^{t-1}\m_{j}\right)\geq 0.$ 
   \STATE Set $\tau=\tau+1$, $l_{\tau}=t$
   \ENDIF
   \ENDFOR
\end{algorithmic}
\end{algorithm}
We first describe the adversary's policy for obtaining the lower bound. Following previous work \cite{Minimax:Online}, our proposed policy is based on the \emph{linear game}, i.e., in each round, the adversary chooses from a set of bounded linear functions:
$$F=\{f(\cdot):\D\mapsto \R|f(\w)=\m^{\top}\w,\|\m\|=G\},$$
which is a subset of convex functions satisfying Assumptions \ref{ass:1} and \ref{ass:2}. For this setting, the regret can be written as
\begin{equation}
\begin{split}
\label{eqn:section31}
R=\sum_{t=1}^Tf_t(\w_t)-\min\limits_{\w\in\D}\sum_{t=1}^Tf_t(\w)=&\sum_{t=1}^T \m_t^{\top}\w_t-\min\limits_{\w\in\D}\left(\sum_{t=1}^T\m_t\right)^{\top}\w\\
=&\sum_{t=1}^T \m_t^{\top}\w_t+\frac{D}{2}\left\|\sum_{t=1}^T\m_t\right\|,\\
\end{split}
\end{equation}
where the third equality is because the minimum is only obtained when $$\w=-D\frac{\sum_{t=1}^T\m_t}{2\left\|\sum_{t=1}^T\m_t\right\|}.$$ According to \eqref{eqn:section31}, to get a tight lower bound for $R$, we have to make both $\sum_{t=1}^T\m_t^{\top}\w_t$ and $\left\|\sum_{t=1}^T\m_t\right\|$ as large as possible. One classical way to achieve this goal is through the \emph{orthogonal technique} \citep{Minimax:Online,chen2019minimax}, that is, in round $t$, the adversary chooses $\m_t$ such that $\m_t^{\top}\w_t\geq0$ and $\m^{\top}_t(\sum_{i=1}^{t-1}\m_i)\geq0$. Note that such a $\m_t$ can always be found for $d\geq 2$. For this technique, it can be easily shown that $\sum_{t=1}^T\m_t^{\top}\w_t\geq0$, while $\left\|\sum_{t=1}^T\m_t\right\|\geq G\sqrt{T}$, which implies an $\Omega(DG\sqrt{T})$ lower bound.

The above policy does not take the constraint on the player's action shifts into account. In the following, we show that, by designing a more adaptive adversary which automatically adjusts its action based on the player's historical switching costs, we can obtain a tighter lower bound when $S$ is small. The details is summarized in Algorithm \ref{alg:AP}. Specifically, in the first round, after observing the player's action $\w_1$, the adversary just simply chooses $f_1(\w)=\m_1^{\top}\w$ such that $\m_1^{\top}\w_1=0$ (Step 3). For round $t\geq2$, the adversary divides the time horizon into several epochs. Let the number of epochs be $N$. For each round $t$ in epoch $\tau\in[N]$, after obtaining $\w_t$,
the adversary checks if the cumulative switching cost of the player inside epoch $\tau$ exceeds a threshold (Step 6). To be more specific, the adversary will check if
$$ \sum_{j=l_{\tau}+1}^t\|\w_j-\w_{j-1}\|\leq \frac{c}{S},$$
where $l_{\tau}$ is the start point of epoch $\tau$, $\ell_1=1$, and $c>0$ is a constant factor. If the inequality holds, then the adversary will keep the action unchanged (Step 7); otherwise, the adversary will find a new $\m_t$ based on the orthogonal technique, i.e., find $\m_t$ such that $\m_t^{\top}\w_t\geq 0$ and $\m_t^{\top}M_{t-1}\geq0$, where $M_{t-1}=\sum_{j=1}^{t-1} \m_j$, and then start a new epoch (Steps 9-10). 

The essential idea behind the above policy is that the adversary adaptively divides $T$ iterations into $N$  epochs, such that for each epoch $\tau\in[N]$, the cumulative switching cost inside of $\tau$ is upper bounded by $$\sum^{l_{\tau+1}-1}_{j=l_{\tau}+1}\|\w_j-\w_{j-1} \|\leq \frac{c}{S},$$
and for each epoch $\tau\in[N-1]$,
$$\sum^{l_{\tau+1}}_{j=l_{\tau}+1}\|\w_j-\w_{j-1} \|>\frac{c}{S}.$$
The above two inequalities help us obtain novel lower bounds for the two terms at the R.H.S.~of \eqref{eqn:section31} respectively which depend on $S$. Specifically, we prove the following two lemmas.
\begin{lemma}
\label{lemma:1}
We have
$$\sum_{t=1}^T\m_t^{\top}\w_t\geq -\frac{cGT}{S}.$$
\end{lemma}
\begin{lemma}
\label{lemma:2}
We have
$$\left\|\sum_{t=1}^T\m_t\right\|\geq G\frac{T\sqrt{c}}{\sqrt{S^2+c}}.$$
\end{lemma}
By appropriately tuning the parameter $c$, we finally prove the following lower bound.
\begin{theorem}
\label{thm:convex:lowerbound}
For any online algorithm, under any given switching cost budget $S$, Algorithm \ref{alg:AP} can generate a series of loss functions $f_1(\cdot),\dots,f_T(\cdot)$ satisfying Assumptions \ref{ass:1} and \ref{ass:2}, such that
\begin{equation*}
R\geq\\
\begin{cases}
0.5DG\sqrt{T},& S\in[D\sqrt{T},DT]\\
0.05DG\frac{DT}{S},& S\in[D,D\sqrt{T})\\
0.05DGT, & S \in[0,D).
\end{cases}
\end{equation*}
\end{theorem}
\paragraph{Remark} The above theorem implies that, when $S\leq D$, the lower bound for OCO-CSC is linear with respect to $T$; When $S\in[D,D\sqrt{T})$, it's possible to achieve sublinear results, and the lower bound decreases with $T/S$; for sufficiently large $S$, i.e., when $S=\Omega(D\sqrt{T})$, the lower bound is $\Omega(DG\sqrt{T})$, which matches the lower bound for the general OCO problem \citep{Minimax:Online}. Note that in this case the lower bound will not further improve as $S$ increases, which is very different from the switching-constrained setting, where the lower bound is $\Omega(T/\sqrt{K})$, which means that increasing the budget $K$ is always beneficial. 
 
\subsection{Upper Bounds}
In this section, we provide the algorithm for obtaining the upper bound. Before introducing our method, we note that, as mentioned in Section \ref{2.3}, the mini-batch OGD algorithm proposed by \citet{chen2019minimax} enjoys an $O(T/\sqrt{S})$ regret bound for OCO-CSC, which is suboptimal based on the lower bound we constructed at the last section. In the following, we show that, perhaps a bit surprisingly, the classical online gradient descent with an appropriately chosen step size is sufficient for obtaining the matching upper bound. Specifically, in round $t$, we update $\w_t$ by
\begin{equation}
\label{alg:convex}
\w_{t+1}=\Pi_{\D}\left[\w_t-\eta \nabla f_t(\w_t) \right],
\end{equation}
where $\Pi_{\D}[\p]$ denotes projecting $\p$ into $\D$, i.e.,
$$\Pi_{\D}[\p]=\argmin\limits_{\w\in\D}(\w-\p)^{\top}(\w-\p).$$
For this algorithm, we prove the following theoretical guarantee.

\begin{figure}
  \centering
  \includegraphics[width=9.5cm]{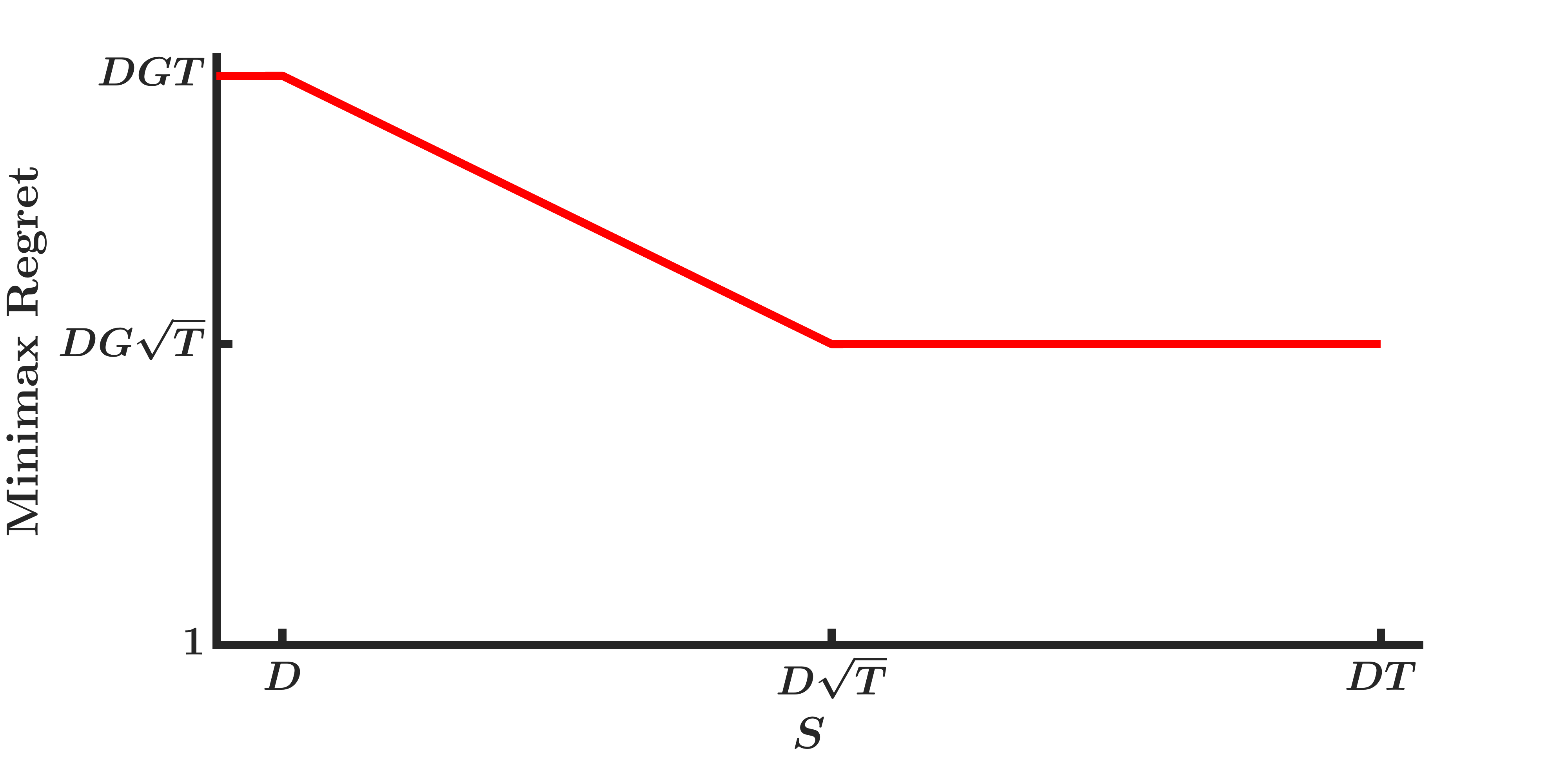}
  \caption{Minimax regret of OCO-CSC. Axies are plotted in log-log scale. }
  \label{fig:1111}
\end{figure}

\begin{theorem}
\label{thm:convex}
Suppose Assumptions \ref{ass:1} and \ref{ass:2} hold, and all loss functions are convex. Then,  under any given switching cost budget $S$, OGD with step size
\begin{equation}
\label{eqn:thm:convex}
\eta=
\begin{cases}
\frac{D}{G\sqrt{T}}, & S\in[D\sqrt{T},DT]\\
\frac{S}{GT},& S\in[0,D\sqrt{T})
\end{cases}
\end{equation}
satisfies \eqref{eqn:intro:ours}, and 
achieves the following regret:
\begin{equation*}
R\leq\\
\begin{cases}
DG\sqrt{T},& S\in[D\sqrt{T},DT]\\
DG\frac{DT}{S},& S\in[D,D\sqrt{T})\\
DGT, & S \in[0,D).
\end{cases}
\end{equation*}
\end{theorem}
\paragraph{Remark} Theorems \ref{thm:convex:lowerbound} and \ref{thm:convex} show that our proposed algorithm enjoys an $O(GD\sqrt{T})$ minimax regret bound for $S=\Omega(D\sqrt{T})$, and $O(DG\min\{DT/S,T\})$ regret bound for $S=O(D\sqrt{T})$. We illustrate the relationship between the minimax regret and $S$ in Figure \ref{fig:1111}. 

 Although the analysis above implies that the theoretical guarantee for OCO-CSC is unimproveable in general, in the following, we show that tighter bounds is still achievable when the loss functions are strongly convex. Specifically, when there are no constraints on the switching cost, the state-of-the-art algorithm is OGD with a time variant step size $\eta_t=1/[\lambda t]$, which enjoys an $O(\log T)$ regret bound. For the OCO-CSC setting, in order to control the overall switching cost, we propose to add a tuning parameter at the denominator of the step size. To be more specific, in round $t$, we update $\w_t$ by  
\begin{equation}
\label{alg:sc-convex}
\w_{t+1}=\Pi_{\D}\left[\w_t-\eta_t \nabla f_t(\w_t) \right],
\end{equation}
where $\eta_t=\frac{1}{\lambda(t+c)}$, and $c>0$ is a constant factor. By configuring $c$ properly, we can obtain the following regret bound.

\begin{theorem}
\label{thm:sc-convex}
Suppose Assumptions \ref{ass:1} and \ref{ass:2} hold, and all loss functions are $\lambda$-strongly convex. Then, under any given switching cost budget $S$, the algorithm in \eqref{alg:sc-convex} with
\begin{equation}
\label{eqn:thm:sc-convex}
c=
\begin{cases}
0, & S\in[\frac{2G}{\lambda}\log(T+1),DT]\\
\frac{T}{\exp(\frac{\lambda}{2G}S)-1}-1,& S\in[0,\frac{2G}{\lambda}\log(T+1))
\end{cases}
\end{equation}
satisfies \eqref{eqn:intro:ours}, and achieves 
$$R\leq \lambda D^2 + \frac{2G^2}{\lambda}\log\left(T+1\right)$$
for $S\in[\frac{2G}{\lambda}\log(T+1),DT]$, and 
$$R\leq \min\left\{\frac{\lambda TD^2}{\exp(\frac{\lambda}{2G}S)-1}+GS,DGT\right\}$$
for $S\in[0,\frac{2G}{\lambda}\log(T+1))$.
\end{theorem}
\paragraph{Remark} Theorem \ref{thm:sc-convex} implies that, when $S\geq \frac{2G}{\lambda}\log (T+1)$, the proposed algorithm enjoys an $O(\log T)$ optimal regret bound; for $S\leq\frac{2G}{\lambda}\log (T+1)$, the proposed algorithm achieves an $O(T/\exp(S)+S)$ regret bound. To obtain a sublinear regret bound, consider $S=\frac{2G}{\lambda}\log(T^{\alpha}+1)$. In this case, we have 
$$R\leq \lambda D^2 T^{1-\alpha}+\frac{2G^2}{\lambda}\log(T^{\alpha}+1),$$
which is sublinar for $\alpha\in(0,1]$.


\section{Theoretical Analysis}
In this section, we present the proofs for the main conclusions.
\subsection{Proof of Lemma \ref{lemma:1}}
\begin{figure}[t]
  \centering
  \includegraphics[width=7cm]{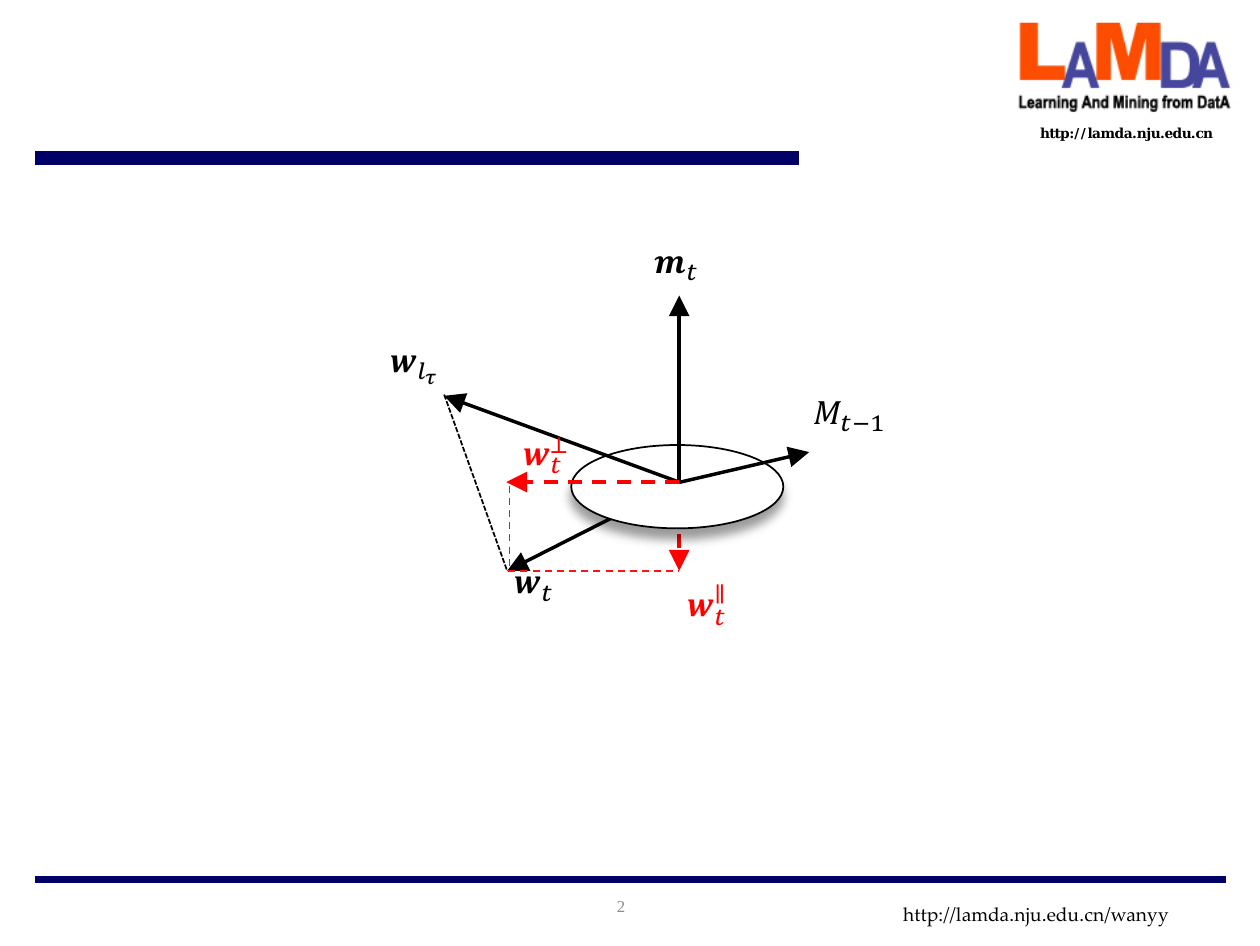}
  \caption{Decomposition of $\w_t$ when $d=3$.}
  \label{fig:decom}
\end{figure}

For any epoch $\tau\in[N]$ of length 1, we have
\begin{equation}
\label{eqn:prove:c:1}
\m_{l_\tau}^{\top}\w_{l_{\tau}}\geq0.
\end{equation}
For any epoch $\tau\in[N]$ whose length is greater than 1, we have $\forall t\in[l_{\tau},l_{\tau+1}-1]$,
\begin{equation}
\label{ineq:proof:lower bound1}
\|\w_t-\w_{l_{\tau}}\|\leq \sum^t_{j=l_{\tau}+1}\|\w_j-\w_{j-1} \|\leq \frac{c}{S},
\end{equation}
where the first inequality is based on the triangle inequality, and the second inequality is guaranteed by Step 6 of the adversary's policy. Next, we decompose $\w_t$ into two terms:  $\w_t=\w_{t}^{\parallel}+\w_{t}^{\perp}$, where
$\w_{t}^{\parallel}$ is the component parallel to $\m_t$, and $\w_{t}^{\perp}$
the component parallel to the  $(d-1)$-normal-hyperplane of $\m_t$. We illustrate the decomposition for $d=3$ in Figure \ref{fig:decom}. Based on the decomposition, we have
\begin{equation}
\begin{split}
\label{eqn:prove:c:2}
\m_{t}^{\top}\w_t= \m^{\top}_{t}(\w_{t}^{\perp}+\w_{t}^{\parallel})= \m_{t}^{\top}\w_{t}^{\parallel}= -G\|\w_{t}^{\parallel}\|\geq  -G\|\w_t-\w_{l_{\tau}}\|\geq  -\frac{cG}{S},
\end{split}
\end{equation}
where the first inequality is based on the triangle inequality, and the fact that $\w_{l_{\tau}}$ is always above the $(d-1)$-normal-hyperplane. The second inequality is derived from \eqref{ineq:proof:lower bound1}.  Combining \eqref{eqn:prove:c:1} and \eqref{eqn:prove:c:2}, we know that for $t\in[T]$,
$$\m_t^{\top}\w_t\geq -\frac{cG}{S},$$
thus
\begin{equation}
\label{eqn:a1}
\sum_{t=1}^T\m_t^{\top}\w_t\geq -\frac{cGT}{S}.
\end{equation}
\subsection{Proof of Lemma \ref{lemma:2}}
Based on Step 6 of the adversary's policy, we know that, for epoch $\tau\in[N-1]$,
\begin{equation}
\sum^{l_{\tau+1}}_{j=l_{\tau}+1}\|\w_j-\w_{j-1} \|>\frac{c}{S},
\end{equation}
since otherwise the adversary will not start a new epoch at $\ell_{\tau+1}$ (note that the cumulative switching cost at the last epoch does not have this lower bound). Thus, the total switching cost in the first $[N-1]$ epochs is lower bounded by
$$\sum_{\tau=1}^{N-1}\sum_{j=l_{\tau}+1}^{l_{\tau+1}}\|\w_j-\w_{j-1}\|>(N-1)\frac{c}{S}.$$
On the other hand, since the overall budget is $S$, we know
$$\sum_{\tau=1}^{N-1}\sum_{j=l_{\tau}+1}^{l_{\tau+1}}\|\w_j-\w_{j-1}\|\leq S.$$
Thus
\begin{equation}
\label{eqn:N}
	N\leq \frac{S^2}{c}+1.
\end{equation}
Let $L_{\tau}$ be the length of epoch $\tau\in[N]$. Based on the Step 9, we know that for each epoch $\tau$, $\m_{l_{\tau}}M_{l_{\tau}-1}\geq0$. Thus, we have
\begin{equation}
\begin{split}
\label{eqn:18}
\left\|\sum_{t=1}^T \m_t \right\|^2=  \left\|\sum_{\tau=1}^N L_{\tau} \m_{l_{\tau}}\right\|^2= \left\| L_{N}\m_{l_{N}}+\sum_{\tau=1}^{N-1} L_{\tau}\m_{l_{\tau}}\right\|^2\geq & \left\| L_{N}\m_{l_{N}}\right\|^2+\left\|\sum_{i=1}^{N-1} L_{\tau}\m_{l_{\tau}}\right\|^2\\
\geq &  G^2\sum_{\tau=1}^N L_i^2.
\end{split}
\end{equation}
Thus
\begin{equation}
\begin{split}
\label{eqn:a2}
\left\|\sum_{t=1}^T\m_t\right\|
\geq  G\sqrt{\sum_{\tau=1}^NL_{\tau}^2}
\geq  G\frac{T}{\sqrt{N}}
\geq  G\frac{T\sqrt{c}}{\sqrt{S^2+c}},
\end{split}
\end{equation}
where the first inequality is derived from \eqref{eqn:18}, the second inequality is based on Cauchy-Schwarz inequality, and the final inequality is derived from \eqref{eqn:N}.
\subsection{Proof of Theorem \ref{thm:convex:lowerbound}}
When $S\geq D\sqrt{T}$, the lower bound can be directly obtained by using the minimax linear game provided by \citep{Minimax:Online}. When $S\in[D,D\sqrt{T})$, by the definition of regret, we have
$$R=\underbrace{\sum_{t=1}^T \m_t^{\top}\w_t}_{{a_1}}\underbrace{-\min\limits_{\w\in\D}\left(\sum_{t=1}^T\m_t\right)^{\top}\w}_{{a_2}}.$$
Based on Lemmas \ref{lemma:1} and \ref{lemma:2}, we get
\begin{equation}
\label{eqn:prove:c:3}
R= a_1+a_2\geq  G\frac{0.5DT\sqrt{c}}{\sqrt{S^2+c}}-\frac{cGT}{S}.
\end{equation}
Let $c=c'D^2$, and we have
\begin{equation*}
\begin{split}
R\geq  G\frac{0.5DT\sqrt{c'D^2}}{\sqrt{S^2+c'D^2}}-\frac{c'D^2GT}{S}= &  GD^2T\left(\frac{0.5\sqrt{c'}}{\sqrt{S^2+c'D^2}}-\frac{c'}{S}\right)\\
\geq & GD^2\frac{T}{S} \left(\frac{0.5\sqrt{c'}}{\sqrt{1+c'}}-c'\right).
\end{split}
\end{equation*}
where the second inequality is due to $D\leq S$. Note that the R.H.S.~of the above inequality is a function of $c'$. To maximize the lower bound, we should solve the following convex problem:
$$\argmax\limits_{x>0} \frac{0.5\sqrt{x}}{\sqrt{1+x}}-x,$$
which is equivalent to finding the solution of the following equation:
$$16x^4+32x^3+49x^2+15x-1=0.$$
it can be easily shown that the optimal solution $x_*\approx 0.056$. Thus, by setting $c'=0.056$, we get
\begin{equation}
R\geq 0.05GD^2\frac{T}{S}.
\end{equation}
 For $S\in(0,D]$, based on \eqref{eqn:prove:c:3} and setting $c=0.056S^2$, we have
\begin{equation*}
\begin{split}
R\geq  GT\left( \frac{0.5D\sqrt{0.056S^2}}{\sqrt{S^2+0.056S^2}}-\frac{0.056S^2}{S} \right)\geq  GDT\left(\frac{0.028}{\sqrt{1.056}}-0.056\right)\geq  0.05GDT.
\end{split}
\end{equation*}
where the second inequality is because $S\leq D$.

\subsection{Proof of Theorem \ref{thm:convex}}
We first prove that by setting $\eta$ as in \eqref{eqn:thm:convex}, the constraint in \eqref{eqn:intro:ours} always holds. Let $\w'_t=\w_{t-1}-\eta\nabla f_{t-1}(\w_{t-1})$. We have
\begin{equation}
\begin{split}
\label{ineq:proof:thm1}
\sum_{t=2}^T\|\w_t-\w_{t-1}\|\leq  \sum_{t=2}^T\|\w'_t-\w_{t-1}\|\overset{\eqref{alg:convex}}{=} \sum_{t=2}^T\|\eta \nabla f_{t-1}(\w_{t-1})\|\overset{\eqref{eqn:gradient}}{\leq} \eta GT,
\end{split}
\end{equation}
where the first inequality is based the following lemma, which describes the non-expansion property of the projection.
\begin{lemma} \emph{\citep{mcmahan2010adaptive}}
\label{lemma:3}
For the projection operation, we have $\forall \w_1,\w_2\in\R^d$,
$$\|\Pi_{\D}(\w_1)-\Pi_{\D}(\w_2)\|\leq \|\w_1-\w_2\|.$$
\end{lemma}
Based on \eqref{ineq:proof:thm1}, for $S\in[D\sqrt{T},DT]$, we have
\begin{equation*}
\begin{split}
\sum_{t=2}^T\|\w_t-\w_{t-1}\|\leq \eta GT
= D\sqrt{T}
\leq & S.
\end{split}
\end{equation*}
For $S\in[0,D\sqrt{T})$, we have
\begin{equation*}
\begin{split}
\sum_{t=2}^T\|\w_t-\w_{t-1}\|\leq \eta GT
= \frac{S}{GT}GT
= S.
\end{split}
\end{equation*}

 Next, we turn to upper bound the regret. Let $\w_*=\argmin_{\w\in\D}\sum_{t=1}^Tf_t(\w)$. Based on the classical analysis of OGD \citep{Intro:Online:Convex}, we have
\begin{equation*}
\begin{split}
\|\w_t-\w_*\|^2\leq \|\w_t'-\w_*\|^2  =& \|\w_{t-1}-\eta\nabla f_{t-1}(\w_{t-1})-\w_*\|^2\\
=&\|\w_{t-1}-\w_*\|^2+\eta^2\|\nabla f_{t-1}(\w_{t-1})\|^2\\
&-2\eta (\w_{t-1}-\w_*)^{\top}\nabla f_{t-1}(\w_{t-1}).
\end{split}
\end{equation*}
Thus
\begin{equation*}
\begin{split}
(\w_t-\w_*)^{\top}\nabla f_t(\w_t)
\leq &\frac{\|\w_{t}-\w_*\|^2-\|\w_{t+1}-\w_*\|^2}{2\eta}+\frac{\eta}{2} \|\nabla f_t(\w_t)\|^2.
\end{split}
\end{equation*}
Based on the convexity of the loss functions and summing the above inequality up from 1 to $T$, we get
\begin{equation}
\begin{split}
\label{eqn:R:ana:cite}
R\leq & \sum_{t=1}^T \frac{\| \w_t-\w_*\|^2-\|\w_{t+1}-\w_*\|^2}{2\eta}+\frac{\eta }{2}\sum_{t=1}^T\|\nabla f_t(\w_t)\|^2\leq  \frac{D^2}{2\eta}+\frac{\eta G^2T}{2}.
\end{split}
\end{equation}
Thus, for $S\in[D\sqrt{T},DT]$, we have
\begin{equation}
\label{eqn:R:cite}
R\leq  \frac{D^2}{2\eta}+\frac{\eta G^2T}{2}=DG\sqrt{T}.
\end{equation}
For $S\in[0,D\sqrt{T})$,
\begin{equation*}
\begin{split}
R\leq  \frac{D^2}{2\eta}+\frac{\eta G^2T}{2}=&DG\left(\frac{DT}{2S}+\frac{S}{2D}\right)\leq DG\left(\frac{DT}{2S}+\frac{\sqrt{T}}{2}\right)\leq DG\left(\frac{DT}{2S}+\frac{DT}{2S}\right)
=DG\frac{DT}{S}.
\end{split}
\end{equation*}
Finally, note that by Assumptions 1 and 2 and the convexity of the loss functions, we always have $R\leq DGT$.

\subsection{Proof of Theorem \ref{thm:sc-convex}}
Let $\w_t'=\w_{t-1}-\eta_{t-1}\nabla f_{t-1}(\w_{t-1}).$ We have
\begin{equation}
\begin{split}
	 \sum_{t=2}^{T}\|\w_t-\w_{t-1}\|
	\leq  \sum_{t=2}^T\|\w_t'-\w_{t-1}\|
	\leq  \frac{G}{\lambda}\sum_{t=1}^T \frac{1}{t+c}.
\end{split}
\end{equation}
where the first inequality is based on Lemma \ref{lemma:3}. To further upper bound the theorem, we introduce the following lemma.
\begin{lemma}\emph{\citep{gaillard2014second}}
Let $a_0>0$ and $a_1,\dots,a_m\in[0,1]$ be real numbers and let $f:[0,+\infty)\mapsto[0,+\infty)$ be a nonincreasing function. then
\begin{equation*}
	\sum_{i=2}^ma_if(a_0+\dots+a_{i-1})\leq f(a_0)+\int_{a_0}^{a_0+\dots+a_m} f(x)dx.
\end{equation*} 
\end{lemma}
Based on the lemma above, we have 
\begin{equation}
	\begin{split}
	\sum_{t=2}^{T}\|\w_t-\w_{t-1}\|
			\leq & \frac{G}{\lambda}\frac{1}{1+c}+\frac{G}{\lambda}\left(\log(T+c)-\log(1+c)\right)\\
	\leq & \frac{G}{\lambda}\frac{1}{1+c}+\frac{G}{\lambda}\log\left(\frac{T}{1+c}+1\right)
	\leq  \frac{2G}{\lambda}\log\left(\frac{T}{1+c}+1\right),
	\end{split}
\end{equation}
where the last inequality is because $1/x\leq \log(T/x+1)$ for any $x\geq1$ and $T\geq3$.
Thus, for $S\geq\frac{2G}{\lambda}\log(T+1)$, by setting $c=0$, we have
$$\sum_{t=2}^T \|\w_t-\w_{t-1}\|\leq \frac{2G}{\lambda}\log\left(\frac{T}{1+c}+1\right)\leq S.$$
When $S\leq \frac{2G}{\lambda}\log(T+1)$, we configure $$c=\frac{T}{\exp(\frac{\lambda}{2G}S)-1}-1\geq0,$$ and get $$\sum_{t=1}^T\|\w_{t}-\w_{t-1}\|\leq S. $$

Next, we consider the regret bound. For $t\in[1,T]$, We have 
\begin{equation}
	\begin{split}
		\|\w_{t+1}-\w_{*}\|^2
		\leq & \|\w_{t+1}'-\w_*\|^2\\
		= & \|\w_{t}-\w_*\|^2-2\eta_{t}(\w_{t}-\w_*)^{\top}\nabla f_{t}(\w_{t})
		+\eta_{t}^2\|\nabla f_{t}(\w_{t})\|^2,
	\end{split}
\end{equation}
thus
\begin{equation}
	\begin{split}
		(\w_{t}-\w_*)^{\top}\nabla f_t(\w_t)\leq &\frac{\|\w_t-\w_{*}\|^2-\|\w_{t+1}-\w_t\|^2}{2\eta_t}
		+\frac{\eta_t}{2}\|\nabla f_t(\w_t)\|^2.
	\end{split}
\end{equation}
By the definition of regret and strong convexity, we have
\begin{equation*}
	\begin{split}
		R= &\sum_{t=1}^T f_t(\w_t)-\sum_{t=1}^T f_t(\w_*)\\
		\leq & \sum_{t=1}^T (\w_t-\w_*)^{\top}\nabla f_t(\w_t)-\frac{\lambda}{2}\sum_{t=1}^T \|\w_{t}-\w_*\|^2\\
		\leq &\frac{1}{2} \sum_{t=2}^T\left(\underbrace{\frac{1}{\eta_t}-\frac{1}{\eta_{t-1}}-\lambda}_{=0}\right)\|\w_t-\w_{*}\|^2+\sum_{t=1}^T\frac{\eta_t}{2}G^2
		+\frac{D^2}{2\eta_1}\\
		\leq & \lambda (c+1)D^2 + \frac{2G^2}{\lambda}\log\left(\frac{T}{1+c}+1\right).
	\end{split}
\end{equation*}
Thus, for $S\geq \frac{2G}{\lambda}\log(T+1)$, we have $c=0$, and thus
$$R\leq \lambda D^2 + \frac{2G^2}{\lambda}\log\left(T+1\right).$$
When $S\leq \frac{2G}{\lambda}\log(T+1)$, we have $c=\frac{T}{\exp(\frac{\lambda}{2G}S)-1}-1$, thus
$$R\leq \frac{\lambda TD^2}{\exp(\frac{\lambda}{2G}S)-1}+GS.$$
Finally, under Assumptions \ref{ass:1} and \ref{ass:2}, we always have $R\leq DGT$.
\section{Conclusion and Future Work}
In this paper, we propose a variant of the classical OCO problem, named OCO with continuous switching constraint, where the player suffers a $\ell_2$-norm switching cost for each action shift, and the overall switching cost is constrained by a given budget $S$. We first propose an adaptive mini-batch policy for the adversary, based on which we prove that the lower bound for this problem is $\Omega(\sqrt{T})$ when $S=\Omega(\sqrt{T})$, and $\Omega(\min\{\frac{T}{S},T\})$ when $S=O(\sqrt{T})$. Next, we demonstrate that OGD with a proper configuration of the step size achieves the minimax optimal regret bound. Finally, for $\lambda$-strongly convex functions, we develop a variant of OGD, which has a tunable parameter at the denominator, and we show that it enjoys an  $O(\log T)$ regret bound when $S=\Omega(\log T)$, and an $O(T/\exp(S)+S)$ regret bound when $S=O(\log T)$.

In the future, we would like to investigate how to extend our setting to other online learning scenarios, such as  bandit convex optimization \citep{flaxman2005online}, and OCO in changing environments \citep{hazan2009efficient}. Moreover, it is also an interesting question to study the switching constraint problem under other distance metrics such as the Bregman divergence. 
\bibliography{ref}

\end{document}